%% file: iclr2024_conference.tex
\documentclass{article} %
\usepackage{iclr2024_conference,times}

\input{math_commands.tex}

\usepackage{hyperref}
\usepackage{url}
\usepackage{booktabs}
\usepackage{graphicx}
\usepackage{algorithm}
\usepackage{algpseudocode}
\usepackage[labelfont=bf,font=small]{caption}
\usepackage{wrapfig}
\usepackage{listings}
\usepackage{xcolor}
\usepackage{enumitem}
\usepackage{multirow}
\usepackage{makecell}

\usepackage{inconsolata}
\usepackage{subcaption}

\usepackage{basecommon}
\algtext*{EndIf}
\algtext*{EndFor}
\algtext*{EndWhile}
\algtext*{EndFunction}

\let\cite=\citep

\let\temp\rmdefault
\let\rmdefault\temp

\definecolor{codegreen}{rgb}{0,0.6,0}
\definecolor{codegray}{rgb}{0.5,0.5,0.5}
\definecolor{codepurple}{rgb}{0.07,0,0.53}
\definecolor{codered}{RGB}{189,41,0}
\definecolor{codecomment}{RGB}{153,153,153}
\definecolor{backcolour}{rgb}{0.96,0.96,0.96}
\definecolor{mygreen}{rgb}{0.0, 0.5, 0.0}
\definecolor{royalblue}{rgb}{0.0, 0.14, 0.4}
\definecolor{egyptianblue}{rgb}{0.06, 0.2, 0.65}
\definecolor{royalazure}{rgb}{0.0, 0.22, 0.66}
\definecolor{portlandorange}{rgb}{1.0, 0.35, 0.21}
\definecolor{saddlebrown}{RGB}{139,69,19}
\definecolor{sienna}{RGB}{183,105,68}
\definecolor{saddlebrown}{RGB}{139,69,19}

\algrenewcommand\algorithmicrequire{\textbf{Input:}}
\algrenewcommand\algorithmicensure{\textbf{Output:}}

\hypersetup{
    colorlinks=true,
    linkcolor=sienna,
    citecolor=egyptianblue,
}

\lstdefinestyle{mystyle}{
    backgroundcolor=\color{backcolour},   
    commentstyle=\color{codegreen},
    keywordstyle=\color{codered},
    numberstyle=\tiny\color{codegray},
    stringstyle=\color{codepurple},
    emph={dequantize},
    emphstyle=\color{codered},    
    basicstyle=\ttfamily\footnotesize,
    breakatwhitespace=false,         
    breaklines=true,                 
    captionpos=b,                    
    keepspaces=true,                 
    numbers=left,                    
    numbersep=5pt,                  
    showspaces=false,                
    showstringspaces=false,
    showtabs=false,   
    morekeywords={>,<,.,;,-,!,=,~},
    tabsize=2
}
\title{LQ-LoRA:   Low-rank Plus Quantized Matrix Decomposition for Efficient Language Model Finetuning}

\author{
$\textbf{Han Guo}^{\dagger \star}$
\hspace{10.0mm} $\textbf{Philip Greengard}^\ddagger$ 
\hspace{10.0mm} $\textbf{Eric P. Xing}^{\dagger \diamond}$
\hspace{10.0mm} $\textbf{Yoon Kim}^{\star}$ \vspace{2mm} \\
\normalfont
\textsuperscript{$\dagger$}Carnegie Mellon University, 
\textsuperscript{$\ddagger$}Columbia University \\
\textsuperscript{$\diamond$}Mohamed bin Zayed University of Artificial Intelligence, Petuum Inc. \\
\textsuperscript{$\star$}Massachusetts Institute of Technology \\
\small{\texttt{hanguo@cs.cmu.edu}}, \,\, \small{\texttt{pg2118@columbia.edu}}, \,\,{\texttt{epxing@cs.cmu.edu}}, \,\,  \small{\texttt{yoonkim@mit.edu}}
}

\iclrfinalcopy %
\begin{document}

\maketitle
\vspace{-4mm}
\begin{abstract}
\vspace{-3mm}
We propose a simple approach for memory-efficient adaptation of pretrained language models. Our approach uses an iterative  algorithm  to decompose each  pretrained  matrix into a high-precision low-rank component  and a memory-efficient quantized component. During finetuning, the quantized component remains fixed and only the  low-rank component is updated. We present an integer linear programming formulation of the quantization component which enables dynamic  configuration of quantization parameters (e.g., bit-width, block size) for each matrix given an overall target memory budget.  We further explore a data-aware version of the algorithm which uses an approximation of the Fisher information matrix to weight the  reconstruction objective during matrix decomposition. Experiments on finetuning RoBERTa and LLaMA-2 (7B and 70B) demonstrate that our low-rank plus quantized matrix decomposition approach (LQ-LoRA) outperforms strong QLoRA and GPTQ-LoRA baselines and enables aggressive quantization to sub-3 bits with only minor performance degradations. When finetuned on a language modeling calibration dataset, LQ-LoRA can also be used for model compression; in this setting our 2.75-bit LLaMA-2-70B model (which has 2.85 bits on average when including the low-rank components and requires 27GB of GPU memory) performs respectably compared to the 16-bit baseline.\footnote{Our code and models are available at \url{https://github.com/HanGuo97/lq-lora}. This work was completed while Han Guo was a visiting student at MIT.}
\vspace{-4mm}

\end{abstract}

\input{introduction}

\input{background}
\input{methods}
\input{experiments}

\input{discussions}

\input{related-works}
\input{conclusion}

\section*{Acknowledgements}
We thank Minyoung Huh, Isha Puri, Hongyi Wang, Lirui Wang, and the members of the Hyundai 42dot research team for helpful comments and discussions. We are also grateful to Mengzhao Chen for clarification questions regarding OmniQuant.
Eric Xing and Han Guo acknowledge the support of Microsoft PhD Fellowship, NGA HM04762010002, NSF IIS1955532, NIGMS R01GM140467, NSF IIS2123952, NSF BCS2040381, NSF IIS2311990, SRC AIHW 2024AH3210, and DARPA ECOLE HR00112390063. This study was supported by funds from Hyundai Motor Group, MIT-IBM Watson AI Lab, and  the MLA@CSAIL initiative.

\bibliography{references}
\bibliographystyle{iclr2024_conference}

\appendix
\input{appendix}

\end{document}

%% file: math_commands.tex
\usepackage{amsmath,amsfonts,bm}

\def\eqref#1{equation~\ref{#1}}

\def\1{\bm{1}}

\DeclareMathAlphabet{\mathsfit}{\encodingdefault}{\sfdefault}{m}{sl}
\SetMathAlphabet{\mathsfit}{bold}{\encodingdefault}{\sfdefault}{bx}{n}

\DeclareMathOperator*{\argmin}{arg\,min}

\DeclareMathOperator{\clamp}{clamp}

%% file: introduction.tex
\vspace{-2mm}
\section{Introduction}
\vspace{-2mm}
Despite the increased availability of large language models (LLMs) and their pretrained parameters \cite{zhang2022opt,lecao2022bloom,touvron2023llama,touvron2023llama2}, their sheer size makes them  expensive to adapt to new datasets via full finetuning. This is particularly unideal since a small amount of  supervised finetuning on instruction following data has been shown to be an effecive approach for learning interactive agents that can follow general instructions  \cite{wang-etal-2023-self-instruct,alpaca,vicuna2023open,zhou2023lima}, and moreover, LLMs finetuned via reinforcement learning with human feedback \cite{ouyang2022instructgpt} represent some of the most capable AI systems that exist today \cite{openai2023gpt4,bubeck2023sparks}. Improving the memory-efficiency of LLM finetuning thus remains a key step in widening the scope of problems to which LLMs can be practically applied. 

One promising framework for memory-efficient LLM adaptation is through parameter-efficient finetuning  methods, which typically learn a smaller finetunable \emph{extension} to the base pretrained model (see \citet{ding2023peft} for a survey). These methods can  reduce the amount of memory required for finetuning  as the pretrained parameters remain fixed---thus reducing the need to allocate memory for storing gradients and optimizer states for these parameters---while the number of new parameters to be optimized is a fraction of the fixed parameters. Of the many existing  parameter-efficient finetuning methods, low-rank adaptation \cite[LoRA;][]{hu2022lora} has emerged as a popular technique for efficient LLM adaptation. In LoRA, the pretrained model's weight matrix $\boldW$ is reparameterized as $\boldW + \boldL_1\boldL_2$,  and only $\boldL_1$ and $\boldL_2$ are finetuned. Recent works have improved the memory-efficiency of LoRA further by applying it to a  quantized pretrained model, i.e., using the reparameterization $q(\boldW) + \boldL_1\boldL_2$ where $q(\cdot)$ is some quantization function \cite{dettmers2023qlora,chai2023int2}.

In LoRA, $\boldL_2$ is initialized to $\mathbf{0}$ to ensure that  the model output is the same as the pretrained model at the beginning of finetuning (i.e., $\boldX(\boldW + \boldL_1\boldL_2) = \boldX\boldW$).  However, if the pretrained matrices are quantized to the extent where there is substantial quantization error (which has been empirically found to occur at sub-4-bit regimes), zero initialization may not be optimal since $q(\boldW) + \boldL_1\boldL_2 \ne \boldW$. In this paper, we exploit the fact that LoRA only performs low-rank updates to the quantized model to derive an initialization scheme that takes the quantization error into account. We use an iterative algorithm similar to those used in the robust PCA literature \cite{wright2009robust,candes2011robust,zhou2011godec} to  decompose $\boldW$ such that $\boldW \approx \boldQ + \boldL_1\boldL_2$. Here $\boldQ$ is the quantized component which remains fixed and $\boldL_1\boldL_2$ is the  low-rank component. During adaptation only $\boldL_1$ and $\boldL_2$ (which captures the high-variance  subspaces of $\boldW$) are finetuned. 
Instead of applying the same quantization configuration to all layers, we use integer linear programming to find a mixed quantization strategy that allows for the assignment of different configurations (bits, block size, etc.) to each matrix given an overall target bit rate. Finally, we explore a data-aware version of the algorithm which modifies the decomposition objective with an approximation of the Fisher information matrix obtained from calibration samples.

We apply LQ-LoRA to adapt RoBERTa \cite{liu2019roberta} and LLaMA-2 \cite{touvron2023llama2} models and find that it can meaningfully improve upon strong QLoRA \cite{dettmers2023qlora} and GPTQ-LoRA \cite{frantar-gptq,chai2023int2} baselines while enabling users to flexibly set a target memory budget. LQ-LoRA can also be applied on standard language modeling datasets to serve as a weight-only post-training quantization (PTQ) method. In this setting we find that we are able to compress LLaMA-2-70B to 2.85 bits with only a small perplexity degradation.

%% file: background.tex
\vspace{-2mm}
\section{Background}
\vspace{-2mm}
\subsection{Low-rank Adaptation of Large Language Models}
\vspace{-2mm}
\label{subsec:lora}
Low-rank adaptation of large language models \cite[LoRA;][]{hu2022lora} has emerged as a simple but effective approach for reducing the memory footprint during LLM finetuning. Given a  matrix $\boldW \in \reals^{d \times k}$ of a pretrained linear layer, LoRA initializes two matrices $\boldL_1 \in \reals^{d \times r}, \boldL_2 \in \reals^{r \times k}$ with $r < \min(d,k)$,   where $\boldL_1$ is initialized to Gaussian noise and $\boldL_2$ is initialized to $\mathbf{0}$ (in order to ensure that $\boldL_1\boldL_2 = \mathbf{0}$ at the start of training).  LoRA then reparameterizes the linear layer as $\boldX(\boldW + \boldL_1\boldL_2)$ (here $\boldX$ is the previous layer's activation), and only finetunes  $\boldL_1$ and $\boldL_2$ during language model adaptation. (The bias vector is omitted for brevity.)

LoRA is more memory-efficient  than  full finetuning as there is no need to allocate GPU memory for the gradients and the associated optimizer states  (e.g., the momentum and variance statistics in Adam \cite{kingma2015adam}) for $\boldW$. Perhaps more so that other strategies for memory-efficient finetuning which also learn a small number of parameters on top of the pretrained model (e.g., Adapters \cite{houlsby2019parameter} and Prompt Tuning \cite{li-liang-2021-prefix,lester-etal-2021-power}), LoRA  has become popular for adapting LLMs, especially for supervised finetuning on instruction-following benchmarks.

\vspace{-2mm}
\subsection{Weight Quantization of Large Language Models}
\vspace{-2mm}
\label{subsec:nf}

Standard round-to-nearest (RTN) quantization, which quantizes/dequantizes a block of weights as $\boldu \approx s \times \clamp \left( \left\lfloor \frac{1}{s}  \boldu \right\rceil; -2^{b-1}, 2^{b-1} - 1\right)$ with scaling factor $s = \frac{\max(|\boldu|)}{2^{b-1}-1}$ and bit size $b$,  has been shown to be effective for quantizing a pretrained LLM's weights to  $8$-bits \cite{yao2022zeroquant}. However, (sub) $4$-bit quantization has been empirically found to be difficult with RTN, and recent methods generally employ a data-aware strategy  which uses calibration samples to obtain better weight quantization  \cite[][\textit{inter alia}]{frantar-gptq,dettmers2022llm8,xiao2022smoothquant,kim2023squeezellm,lin2023awq,dettmers2023spqr,shao2023omniquant}.

Our approach relies on the recently proposed   {NormalFloat} (NF) quantization scheme \cite{dettmers2023qlora}, which exploits the fact that the distribution of the weights of a trained model is approximately Gaussian.  Following the presentation from \citet{yoshida2023nf4}, NF quantization  calculates $2^{b-1}$ evenly-spaced  values from $[\delta, \frac{1}{2}]$, and $2^{b-1}+1$ evenly-spaced values from $[\frac{1}{2}, 1-\delta]$, where $\delta = \frac{1}{2}(\frac{1}{30} + \frac{1}{32})$. This results in $2^{b}$ probability values $[p_1, \dots, p_{2^{b}}]$ where $p_1 = \delta,  p_{2^{b-1}} = \frac{1}{2}$, and $p_{2^{b}} = 1-\delta$. These probabilities are converted into quantiles $[q_1, \dots, q_{2^{b}}]$ where $q_i = \Phi^{-1}(p_i)$ is the Gaussian quantile for $p_i$, and these quantiles are normalized to $[-1,1]$ by  $\tilde{q}_{i} = \frac{q_i}{q_{2^{b}}}$. Then, given a block of weights $\boldu=[u_1, \dots, u_B]$ and the absmax value $s = \max(|\boldu|)$ for that block, the weights $u_j$ in this block are quantized to the nearest quantile $c_j$, i.e., 
    $c_{j} = \argmin_{i \in \{1, \dots, 2^{b}\}} \left|\tilde{q_i} - \frac{u_j}{s} \right|$.

For a $d \times k$ matrix there are $\frac{dk}{B}$ blocks, and hence storing the absmax values $s$ for each block could become substantial with small block sizes. \citet{dettmers2023qlora} thus employ a double quantization strategy where the set of absmax values $[s_1, \dots, s_{\frac{dk}{B}}]$ for a given matrix are quantized again via RTN. Based on this quantization scheme, \citet{dettmers2023qlora} propose QLoRA, which performs NF quantization to 4 bits on the pretrained LLM, and learns low-rank updates. QLoRA has been found to be competitive with full finetuning across a number of benchmarks, and thus serves as the main baseline of the present work.

%% file: methods.tex
\vspace{-2mm}
\section{Method: LQ-LoRA}
\vspace{-2mm}
\label{sec:methods}

Our approach relies on a simple factorization scheme which decomposes each pretrained matrix into a low-rank matrix plus a quantized matrix (\S \ref{subsec:lpq}), where only the low-rank component is adapted during finetuning. In \S\ref{subsec:ilp} we explore a mixed quantization strategy via integer linear programming to allow for dynamic quantization across layers given a target average bit rate. We further consider a data-aware version of LQ-LoRA by using the empirical Fisher information matrix to weight the  reconstruction objective during matrix factorization (\S\ref{subsec:fisher}).

\vspace{-2mm}
\subsection{Low-rank Plus Quantized Matrix Decomposition}
\vspace{-2mm}
\label{subsec:lpq}
As noted in \S\ref{subsec:lora}, LoRA reparameterizes a pretrained matrix as $\boldW$ as $\boldW + \boldL_1 \boldL_2$ and initializes $\boldL_1$ from a Gaussian and $\boldL_2$ to $\mathbf{0}$ before finetuning. While this ensures that the model output is exactly the same as before reparameterization at the start of finetuning, it may present an issue when working with a quantized version of $\boldW$ since we could have $\Vert \boldW - \quantize(\boldW) \Vert_{F} \gg 0$ when quantizing to low bits. This initialization moreover does not take into account $\boldW$'s structure  when deciding on which subspaces to adapt. We approach this problem from the perspective of matrix factorization where we are interested factorizing the original matrix into an easily quantizable component and a low-rank component that captures high-variance directions,
\begin{equation}
\begin{aligned}
&\argmin_{\mathbf{Q}, \mathbf{L}_1, \mathbf{L}_2} \, \Vert \boldW - (\boldQ + \boldL_1 \boldL_2) \Vert_F, 
&& \text{where} \,\, \boldQ \in \mathbb{Q}_b^{d \times k}, \boldL_1 \in \reals^{d \times r}, \boldL_2 \in \reals^{r \times k}.
\label{eq:lpq}
\end{aligned}
\end{equation}
Here $\mathbb{Q}_b^{d \times k} \subset \reals^{d \times k}$ is the set of matrices that are losslessly NF-quantizable to $b$-bits. This optimization problem is similar to the one faced in robust principal components analysis \citep[RPCA;][]{wright2009robust,candes2011robust}, which aims to decompose a matrix $\boldW$ into $\boldL + \boldS$ where $\boldL$ is low-rank and $\boldS$ is \emph{sparse}. Following iterative algorithms which have been shown to be effective for RCPA~\citep{lin2010augmented,zhou2011godec}, we approximately solve Eq.~\ref{eq:lpq} via
 alternating between optimizing $\boldL_1 \boldL_2$, and $\boldQ$:\footnote{In practice we use randomized  SVD instead of full SVD, which significantly reduced runtime for the SVD portion of the algorithm without much deterioration in performance.} 
\begin{equation}
\begin{aligned}
    \boldL_1^{(t)}, \boldL_2^{(t)} &\gets \operatorname{SVD}(\boldW - \boldQ^{(t-1)}, r), && \hspace{0.5cm} {\color{gray} = \argmin_{\rank(\boldL) \le r} \, \Vert \boldW - (\boldQ^{(t-1)} + \boldL) \Vert_{F},} \\
    \boldQ^{(t)} &\gets \operatorname{Quantize}(\boldW - \boldL_1^{(t)} \boldL_2^{(t)}), && \hspace{0.5cm} {\color{gray} \approx \argmin_{\boldQ \in \mathbb{Q}^{d \times k}_b} \, \Vert \boldW - (\boldQ + \boldL_1^{(t)}\boldL_2^{(t)}) \Vert_F,}
\end{aligned}
\label{eq:lpq-steps}
\end{equation}
where $\boldQ^{(0)}$ is initialized to $\mathbf{0}$. Unlike (some) RPCA algorithms for which theoretical convergence guarantees can be obtained \cite{ma2018efficient}, the above algorithm is  heuristic. We thus employ a simple stopping criterion where we keep track of the error $\Vert \boldW - (\boldQ^{(t)} + \boldL_1^{(t)} \boldL_2^{(t)}) \Vert_F$ and terminate the algorithm if the error increases. The iterative decomposition algorithm is shown in Algorithm~\ref{alg:lpq}.\footnote{Despite the simplity of our approach, we are not aware of prior work on low-rank plus quantized matrix decomposition, except for a recent preprint which proposes to perform SVD on the residuals $\boldE = \boldW - \quantize(\boldW)$ to correct for errors after quantization \cite{yao2023zeroquant}. This approach can be seen as performing a single step of the iterative algorithm with the initialization $\boldQ^{(0)} = \quantize(\boldW)$. In our experiments we did not observe significant differences in performance when we  initialized $\boldQ^{(0)}$ to $\quantize(\boldW)$.} Each step of the  algorithm (i.e., randomized SVD followed by quantization) takes a few seconds on a modern GPU for a $4096 \times 4096$ matrix.

\begin{figure}[t]
\centering
\vspace{-12mm}
\includegraphics[width=.9\textwidth]{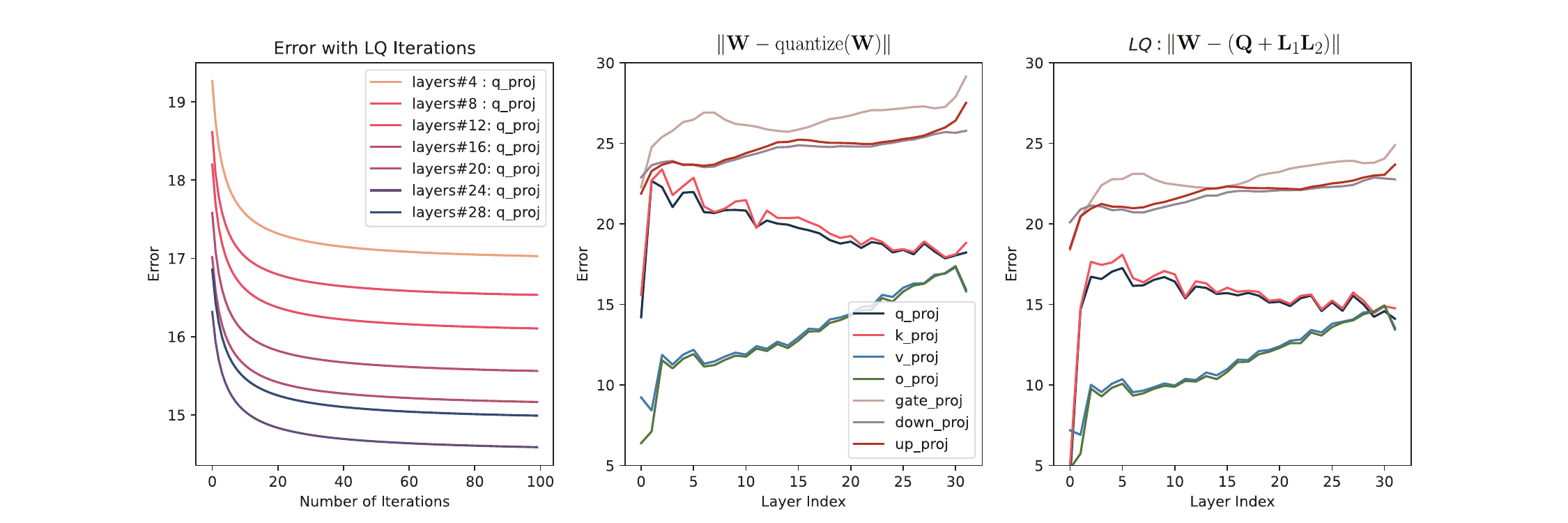}
\vspace{-4mm}
\caption{(Left) The decomposition error $\Vert \boldW - (\boldQ + \boldL_1 \boldL_2) \Vert_F$ for the query projection matrices for different layers of LLaMA-2-7B as a function of the number of LQ steps. (Center) Quantization error for NF-3 quantization for all layers. (Right) LQ decomposition error for 3-bit quantization with rank = 64. LQ decomposition results in less quantization error.}
\label{fig:lpq_error}
\vspace{-5mm}
\end{figure}

 \vspace{-2mm}
\paragraph{Preliminary experiments.}
In Figure~\ref{fig:lpq_error} (left) we show the decomposition error $\Vert \boldW - (\boldQ + \boldL_1 \boldL_2)\Vert_F$ for a few layers of LLaMA-2-7B as a function of the number of steps. We find that our algorithm, while heuristic, is empirically effective. In Figure~\ref{fig:lpq_error} (center) we show the quantization error for 3-bit NF quantization for all matrices, while in Figure~\ref{fig:lpq_error} (right) we show the corresponding error for LQ decomposition. For both approaches we find that the value and output projection matrices become harder to quantize at deeper layers, while the key and query matrices become easier; however, our LQ decomposition is able to  improve upon  vanilla quantization for all layers.

\vspace{-2mm}
\subsection{Mixed-Configuration Quantization via an Integer Linear Program}
\vspace{-2mm}
\label{subsec:ilp}
LQ-LoRA uses the NormalFloat (NF) quantization scheme from \citet{dettmers2023qlora} to quantize the residual $\boldQ$ at each time step.  NF-quantization has several parameters that affect the overall compression rate such as the number of  quantile bins, number of blocks, and bits for double quantization. In this paper we work with slightly different variant which quantizes a matrix $\boldA$ via the following:
\begin{equation*}
\begin{aligned}
\widehat{\boldA}, \bolds =  \operatorname{Quantize-NF}\left(\boldA, b_0, B_0 \right), \,\,\,\,\,\,\,\, &  \widehat{\bolds}, \boldv = \operatorname{Quantize-INT}\left(\bolds, b_1, B_1 \right), && \widehat{\boldv} =  \operatorname{cast}\left( \boldv, b_2 \right).
\end{aligned}
\end{equation*}
Concretely, we first apply NF-quantization with bit size $b_0$ and bucket size $B_0$ to obtain the quantized matrix $\widehat{\boldA}$ and the absmax values for each block $\bolds = [s_1, \dots, s_{\frac{dk}{B_0}}]$ (see \S\ref{subsec:nf}). These absmax values are further quantized to $b_1$ bits via uniform integer quantization with bucket size $B_1$ to obtain the  quantized vector $\widehat{\bolds}$, along with the absmax values for $\bolds$, i.e., ${\boldv} = [v_1, \dots v_{\frac{dk}{B_0 B_1}}]$.\footnote{I.e., given   $v_1 =  \operatorname{absmax}([s_1, \dots, s_{B_1}])$ for a group of size $B_1$ we have $\hat{s}_i = \clamp\left(\lfloor \frac{s_i}{v_1} \rceil; 0, 2^{b_1 - 1}\right)$.} Finally, we cast $\boldv$ to $b_2$ bits to obtain $\widehat{\bolds_1}$.\footnote{This approach deviates from the original approach in that we use  integer quantization on $\bolds$ as opposed to \texttt{FP8} (which did not affect results), and we cast $\boldv$ to lower precision (which led to negligible increase in error).} Dequantization, which is needed on the fly for finetuning and inference, simply reverses this process.

This quantization scheme requires storing $\widehat{\boldA}, \widehat{\bolds}, \widehat{\boldv}$ to represent $\boldA$. We can thus quantify the storage cost (number of bits) for storing $\boldA$ given a configuration $c = (b_0, b_1, b_2, B_0, B_1)$ as 
\begin{equation}
\operatorname{storage}(\boldA, c) = \operatorname{sizeof}(\boldA) \cdot \left( b_0 + \frac{b_1}{B_0} +  \frac{b_2}{B_0 \cdot B_1} \right).
\label{eq:storage}
\end{equation}
The original NF-$4$ double quantization is a special case with $c_{\operatorname{NF4}}=(4,8,\texttt{fp32},64,256)$ and $\operatorname{storage}(\boldA, c_{\operatorname{NF4}}) = 4.127 \cdot \operatorname{sizeof}(\boldA)$, i.e., NF-$4$ requires on average 4.127 bits per parameter.

\input{tables/algorithm}

\begin{wraptable}{r}{0.32\textwidth}
\vspace{-5mm}
\centering
\begin{tabular}{c l}
\toprule
 & Configuration grid \\
\midrule
$b_0$ & $\{2, 3, 4\}$ \\
$b_1$ & $\{2, 3, 4\}$ \\
$b_2$ & $\{\texttt{bf16}, \texttt{fp16}, \texttt{fp32}\}$ \\
$B_0$ & $\{16, 32, 64\}$ \\
$B_1$ & $\{16, 64, 256\}$ \\
\bottomrule
\end{tabular}
\vspace{-2.5mm}
\caption{Search space $\mathcal{C}$ for the  NF quantization configurations.}
\label{tab:search}
\vspace{-7mm}
\end{wraptable}
\vspace{-2mm}
\paragraph{Dynamic quantization configurations.} Prior works on quantizing LLMs have generally focused on applying the same quantization strategy to each matrix, which cannot adapt to users' varying resource constraints and moreover may be suboptimal given that some matrices may be harder to quantize than others. We explore a mixed-precision quantization strategy based on integer linear programming \cite{yao2021hawq,tang2022mixed,kundu2022bmpq}, which allows for the allocation of different configurations to each matrix given a user-defined target target bit rate.

Let $c^{} = (b_0^{}, b_1^{}, b_2^{}, B_0^{}, B_1^{})$ be the configuration parameters and further let $\mathcal{C}$ be the set of possible configurations which is specified by the user (see Table~\ref{tab:search} for the settings we consider in this work). 

Letting $\{ \boldW^{(i)}\}_{i \in [N]}$ be the set of $N$ matrices in an LM, our goal is to find an assignment matrix $\mathbf{X} \in \{0, 1\}^{N \times |\mathcal{C}|}$ that minimizes the Frobenius norm between the matrices before and after low-rank plus quantized decomposition, while respecting a target memory budget. One way to approach this optimization problem is through the following integer linear program,\footnote{Here we overload $c$ to refer to both its tuple representation $c \in \mathcal{C}$ and its index representation $c \in [|\mathcal{C}|].$}
\begin{equation*}
\begin{aligned}
&\min_{\boldX \in \{0, 1\}^{N \times |\mathcal{C}|}} && \sum_{i \in [N]} \sum_{c \in \mathcal{C}} \operatorname{error}(\boldA^{(i)}, c) \cdot \mathbf{X}[i, c], \\
& \text {subject to} && \sum_{i \in [N]} \sum_{c \in \mathcal{C}} \operatorname{storage}(\boldA^{(i)}, c) \cdot \mathbf{X}[i, c] \leq \text{budget}, \\
& &&\sum_{c \in \mathcal{C}} \mathbf{X}[i, c] = 1, \quad \forall i \in [N].
\end{aligned}
\label{eq:ilp}
\end{equation*}
Here $\operatorname{error}(\boldW^{(i)}, c^{}) = \Vert \boldW^{(i)} - (\boldQ + \boldL_1 \boldL_2) \Vert_F^2$ is the reconstruction error after running the iterative algorithm from Sec.~\ref{subsec:lpq} where the $\quantize$ function uses configuration $c$. To approximately solve this ILP we  pre-compute the errors for all matrices and quantization configurations ($|\mathcal{C}| = 3^5$) and use an off-the-shelf solver,\footnote{\url{https://www.gurobi.com/}} as shown in Algorithm~\ref{alg:ilp}. The pre-computation is a one-time process and takes a few hours when parallelized across four A100 GPUs for LLaMA-2-7B. Once the (approximately) optimal configuration $c^{(i)}$ is found, we apply the decomposition on $\boldW^{(i)}$ one more time with $c^{(i)}$ to obtain the final matrices $\boldQ^{(i)}$, $\boldL_1^{(i)}, \boldL^{(i)}_2$ for $i \in [N]$ (see Algorithm~\ref{alg:lplora}).

\vspace{-2mm}
\paragraph{Implementation.}   Existing weight-only quantization implementations often use custom CUDA extensions that are dependent on a particular quantization configuration, making it difficult to extend to mixed-quantization strategies. Our implementation is based  on PyTorch for flexible experimentation and implementation.
We use PyTorch’s \texttt{\_\_torch\_dispatch\_\_} functionality to duck-type \texttt{torch.Tensor}. This allows us to overload PyTorch operations such as matrix multiplication to perform just-in-time dequantization. We then use PyTorch’s (full-graph) compiler to compile the bits-unpacking, dequantization, other linear algebra operations. For batch size $>$ 1, this PyTorch-based implementation (followed by compilation) was  as fast as some custom  CUDA implementations such as \texttt{bitsandbytes}.\footnote{\url{https://github.com/TimDettmers/bitsandbytes}} Further details and speed comparisons are given in Appendix~\ref{subsec:implementation}.

\vspace{-2mm}
\subsection{Data-Aware Matrix Decomposition via Fisher-weighted SVD}
\vspace{-2mm}
\label{subsec:fisher}
The decomposition objective considered in \S\ref{subsec:lpq} is data-agnostic insofar as it treats each entry of $\boldW$ as equally important for reconstruction during factorization. Following recent works which demonstrate the importance of using calibration data for quantizating LLMs \cite{frantar-gptq,lin2023awq,kim2023squeezellm}, we next consider a data-aware version of the approach by using a diagonal approximation of the Fisher information matrix to weight the reconstruction objective. The (diagonal of the) empirical Fisher information matrix for $\boldW$ is  given by $\boldF \in \reals^{d \times k}$ where each entry of the matrix is the averaged square of the derivative over $D$ samples, i.e.,
$
    \boldF_{ij} = \frac{1}{D} \sum_{d=1}^{D} \left( \frac{\partial}{\partial \boldW_{ij}} \log p_{\text{LM}}\left(\boldx^{(d)}\right)\right)^2.
$
Intuitively, this metric measures how sensitive the model's output is to a perturbation of each parameter, and has  previously been exploited to improve  low-rank compression \citep{hsu2021language} and quantization \citep{kim2023squeezellm} of pretrained language models. We similarly use $\boldF$ to weight the decomposition objective, 
\begin{equation}
\begin{aligned}
\left\Vert {\sqrt{\boldF}\, \odot } \left(\boldW - \left(\boldQ + \boldL_1 \boldL_2\right) \right) \right\Vert_F^2,
\end{aligned}
\label{eq:weighted-error}
\end{equation}
where $\odot$ is the Hadamard product. When applied to the LQ decomposition algorithm from \S\ref{subsec:lpq}, this results in the following weighted SVD problem, where given $\boldE := \boldW - \boldQ$ and weighting matrix $\mathbf{F}$, we must find matrices $\mathbf{L}_1 \in \mathbb{R}^{d \times r}, \mathbf{L}_2 \in \mathbb{R}^{r \times k}$ that form the best rank-$r$ approximation,
\begin{equation*}
\begin{aligned}
\mathbf{L}_1, \mathbf{L}_2
= \argmin_{\mathbf{L}_1, \mathbf{L}_2}  \; \left\Vert \sqrt{\mathbf{F}} \odot \left( \boldE  -  \mathbf{L}_1 \mathbf{L}_2\right)\right\Vert_F^2 .
\end{aligned}
\end{equation*}
Unliked its unweighted counterpart, this problem is in general intractable \citep[and in fact NP-hard;][]{razenshteyn2016weighted} and is typically addressed through approximate methods~\citep{srebro2003weighted,li2016recovery,tuzhilina2021weighted}.
However, if we assume that either rows or columns of  the weight matrix $\mathbf{F}$ have identical values, we have the following identity,
\begin{equation*}
\begin{aligned}
\mathbf{L}_1, \mathbf{L}_2
&= \argmin_{\mathbf{L}_1, \mathbf{L}_2}  \; \left\| \sqrt{\mathbf{F}} \odot \left( \boldE -  \mathbf{L}_1 \mathbf{L}_2\right)\right\|_F^2 %
&= \argmin_{\mathbf{L}_1, \mathbf{L}_2}  \; \left\| \mathbf{D}_{\text{row}} \left( \boldE -  \mathbf{L}_1 \mathbf{L}_2\right)\mathbf{D}_{\text{col}}\right\|_F^2, 
\end{aligned}
\end{equation*}
where $\mathbf{D}_{\text{row}}$ is a diagonal matrix consists of row-means of $\sqrt{\mathbf{F}}$, and $\mathbf{D}_{\text{col}}$ is a diagonal matrix consisting of the column-means of $\sqrt{\mathbf{F}}$, i.e., 
\begin{equation*}
\begin{aligned}
\mathbf{D}_{\text{row}} {=}
\operatorname{diag}\left(\left[
\operatorname{avg}(\sqrt{\mathbf{F}_{1, \cdot}}), \dots, \operatorname{avg}(\sqrt{\mathbf{F}_{d, \cdot}})
\right]\right),
\mathbf{D}_{\text{col}} {=}
\operatorname{diag}\left(\left[
\operatorname{avg}(\sqrt{\mathbf{F}_{\cdot, 1}}), \dots,  \operatorname{avg}(\sqrt{\mathbf{F}_{\cdot, k}})
\right]\right).
\end{aligned}
\end{equation*}
In this case the above problem can be solved exactly by standard SVD,
\begin{equation}
\begin{aligned}
\mathbf{U}, \bSigma, \mathbf{V}^\top \gets \operatorname{SVD}(\mathbf{D}_{\text{row}} \boldA \mathbf{D}_{\text{col}}), \,\,\,\,\, & \mathbf{L}_1 \gets \mathbf{D}_{\text{row}}^{-1} \mathbf{U} \sqrt{\bSigma}, && \,\,\,\,\,
\mathbf{L}_2 \gets \sqrt{\bSigma} \mathbf{V}^\top \mathbf{D}_{\text{col}}^{-1}.
\end{aligned}
\label{eq:custom-weighted-svd}
\end{equation}
(See Algorithm~\ref{alg:factorize}.) While the homogenous row/column assumption clearly does not hold for $\boldF$, we found this approach to work well in practice.\footnote{In preliminary experiments we also explored a version of data-aware LQ-LoRA where we approximately minimized $\Vert \boldX(\boldW -  (\boldQ + \boldL_1 \boldL_2))\Vert_F$ using activations $\boldX$ from calibration data, instead of the Fisher information matrix. However we found this to underperform the Fisher approach.} We note that this approximation is a simple extension of~\citet{hsu2021language} who use $\boldD_{\text{row}}$ but not $\boldD_{\text{col}}$ in their weighted SVD (we found that using both the row- and column-averages performed slightly better).

\vspace{-2mm}
\paragraph{Discussion.} This data-aware version of LQ-LoRA requires being able to backpropagate through the pretrained LM in order to obtain the Fisher matrices $\{\boldF^{(i)}\}_{i \in [N]}$, which, in some sense, goes against the setting targeted by memory-efficient adaptation methods wherein full finetuning is not considered possible. This is a valid point, and hence we study both version of LQ-LoRA in our empirical study. We note however, that we compute $\{\boldF^{(i)}\}_{i \in [N]}$ based on some generic text data to obtain the LQ-LoRA initializations  $\{\boldQ^{(i)}, \boldL_1^{(i)}, \boldL_2^{(i)}\}_{i\in [N]}$, and use the \emph{same} initialization for different downstream tasks. This makes the data-aware approach practical, as the Fisher computaton and the matrix decomposition needs to performed only once (as in the non-data-aware version). 

%% file: tables/algorithm.tex
\begin{figure*}[t]
\vspace{-14mm}
\begin{minipage}{0.48\textwidth}

\begin{algorithm}[H]
\footnotesize
\caption{\footnotesize $\operatorname{LQ-LoRA}$ (Section~\ref{sec:methods})}\label{alg:lplora}
\begin{algorithmic}
\Require
$\{\mathbf{\boldW}^{(i)}\}_{i\in[N]}$: Parameters \\
\quad\quad\
$\{\mathbf{\boldF}^{(i)}\}_{i\in[N]}$: \, Fisher information (optional) \\
\quad\quad\
$\mathcal{C}$: List of quantization configurations \\
\quad\quad\
$r$: LoRA rank \\
\quad\quad\
$B_Q$: Quantization budget
\State {\color{gray}\text{\# get quantization configurations (Section~\ref{subsec:ilp}).}}
\State $\{c^{(i)}\} \gets \hspace{-0mm} \operatorname{GetConfig}(\{\mathbf{\boldW}^{(i)}, \mathbf{\boldF}^{(i)}\}, \mathcal{C}, r, B_Q)$
\For{$i \gets 1$ to $N$}
    \State {\color{gray}\text{\# matrix decomposition (Section~\ref{subsec:lpq}).}}
    \State $\mathbf{Q}^{(i)}, \boldL_{1}^{(i)}, \boldL_{2}^{(i)}, \epsilon \gets \operatorname{LQ}(\boldW^{(i)}, \boldF^{(i)}, c^{(i)}, r)$
\EndFor
\Ensure  $\{\mathbf{Q}^{(i)}, \boldL_{1}^{(i)}, \boldL_{2}^{(i)}\}_{i\in[N]}$
\end{algorithmic}
\end{algorithm}%
\vspace{-4mm}
\begin{algorithm}[H]
\footnotesize
\caption{\footnotesize $\operatorname{LQ}$ (Section~\ref{subsec:lpq})}\label{alg:lpq}
\begin{algorithmic}
\Require
$\mathbf{\boldW}$: Input weight matrix \\
\quad\quad\
$\boldF$: Fisher information (optional) \\
\quad\quad\
$c$: Quantization configuration \\
\quad\quad\
$r$: Target rank
\State $\text{Initialize } \mathbf{Q} \gets \mathbf{0}$ and $\epsilon_0 \gets \infty$
\For{$t \gets 1$ to $T$}
    \State $\boldL_{1}, \boldL_{2} \gets \operatorname{Factorize}(\mathbf{W} - \mathbf{Q}, \boldF, r)$
    \State $\mathbf{Q} \gets \operatorname{Quantize}\left({\mathbf{W} - \mathbf{L}_{1} \mathbf{L}_{2}}, c\right)$
    \If{$\mathbf{F}$ is $\operatorname{None}$}
        \State $\epsilon_t \gets \left\Vert\boldW {-} \left(\boldQ {+} \boldL_{1}\boldL_{2}\right)\right\Vert_F$
    \Else
        \State {\color{gray}\text{\# weighted error (Section~\ref{subsec:fisher}).}}
        \State $\epsilon_t \gets \left\Vert\sqrt{\boldF} \odot \left(\boldW {-} \left(\boldQ {+} \boldL_{1}\boldL_{2}\right)\right)\right\Vert_F$
    \EndIf
    \State\algorithmicif\ $\epsilon_t > \epsilon_{t-1}$ \algorithmicthen\ break
\EndFor
\Ensure  $\boldQ, \mathbf{L}_1, \mathbf{L}_2, \epsilon_{t}$
\end{algorithmic}
\end{algorithm}%

\end{minipage}\hfill%
\begin{minipage}{0.5\textwidth}
\small

\begin{algorithm}[H]
\footnotesize
\caption{\footnotesize $\operatorname{GetConfig}$ (Section~\ref{subsec:ilp})}\label{alg:ilp}
\begin{algorithmic}
\Require
$\{\mathbf{\boldW}^{(i)}\}_{i\in[N]}$: Parameters \\
\quad\quad\
$\{\mathbf{\boldF}^{(i)}\}_{i\in[N]}$: \, Fisher information (optional) \\
\quad\quad\
$\mathcal{C}$: List of quantization configurations \\
\quad\quad\
$r$: Target rank \\
\quad\quad\
$B_Q$: Quantization budget
\State $\boldE, \boldS \gets \operatorname{zeros}(N, | \mathcal{C} |)$ {\color{gray}\text{\# initialize error and storage}}
\For{$i \gets 1$ to $N$}
    \For{$c \in \mathcal{C}$}
        \State $\mathbf{Q}^{(i)}, \boldL_{1}^{(i)}, \boldL_{2}^{(i)}, \epsilon \hspace{-1mm} \gets \hspace{-1mm} \operatorname{LQ}(\boldW^{(i)}, \boldF^{(i)}, c, r)$
        \State $\boldE[i, c] \gets \epsilon^2$
        \State $\boldS[i, c] \gets \operatorname{storage}(\boldW^{(i)}, c)$
    \EndFor
\EndFor
\State {\color{gray}\text{\# get optimal configuration given budget with ILP.}}
\State $\{c^{(i)}\}_{i\in[N]} \gets \operatorname{ILPSolve}(\boldS, \boldE, B_Q)$
\Ensure  $\{c^{(i)}\}_{i\in[N]}$
\end{algorithmic}
\end{algorithm}
\vspace{-6.5mm}
\begin{algorithm}[H]
\footnotesize
\caption{\footnotesize $\operatorname{Factorize}$ (Section~\ref{subsec:fisher})}\label{alg:factorize}
\begin{algorithmic}
\Require
$\mathbf{\boldA}$: Input matrix \\
\quad\quad\
$\boldF$: SVD weighting matrix (optional) \\
\quad\quad\
$r$: Target rank
\If{$\boldF$ is $\operatorname{None}$}
    \State {\color{gray}\text{\# (randomized) SVD with target rank} $r$}
    \State $[\mathbf{U}, \mathbf{\Sigma}, \mathbf{V}^\top] \gets \operatorname{SVD}(\boldA, r)$
    \State $\mathbf{L}_1 \gets \mathbf{U} \sqrt{\mathbf{\Sigma}}$, $\,\,\mathbf{L}_2 \gets \sqrt{\mathbf{\Sigma}} \mathbf{V}^\top$
\Else  {\hspace{4mm} \color{gray}\text{\# weighted SVD (Section~\ref{subsec:fisher}).}}
    \State $\mathbf{D}_{\text{row}} \gets \operatorname{RowAverage}(\boldF)$
    \State $\mathbf{D}_{\text{col}} \,\gets \operatorname{ColAverage}(\boldF)$
    \State $[\mathbf{U}, \bSigma, \mathbf{V}^\top] \gets \operatorname{SVD}(\mathbf{D}_{\text{row}} \boldA \mathbf{D}_{\text{col}}, r)$
    \State  $\mathbf{L}_1 \gets \mathbf{D}_{\text{row}}^{-1} \mathbf{U} \sqrt{\bSigma}$, $\,\, \mathbf{L}_2 \gets \sqrt{\bSigma} \mathbf{V}^\top \mathbf{D}_{\text{col}}^{-1}$
    \EndIf
\Ensure  $\mathbf{L}_1, \mathbf{L}_2$
\end{algorithmic}
\end{algorithm}
\end{minipage}
\vspace{-4mm}
\end{figure*}

%% file: experiments.tex
\vspace{-2mm}
\section{Empirical Study}
\vspace{-2mm}
\label{sec:empirical-study}

We conduct experiments with LQ-LoRA across three settings: (1) continual language modeling on C4 training data, (2) instruction tuning on the OpenAssistant dataset~\cite{kopf2023openassistant}, (3) and finetuning on GLUE \cite{wang2018glue}. For (1) and (2) we work with LLaMA-2 models \citep{touvron2023llama2}, while for (3) we use RoBERTa-Large \citep{liu2019roberta}. Our setup closely follows the setup from~\citet{dettmers2023qlora}. The Fisher-weighted version of LQ-LoRA uses randomly sampled sequences from the C4 training set, where for RoBERTa-Large we employ the masked language modeling objective (also on C4) to obtain the Fisher matrix.

\vspace{-2mm}
\paragraph{Baselines.}
Our main baselines include QLoRA~\cite{dettmers2023qlora} and GPTQ-LoRA. Both approaches perform PTQ on the pretrained model before learning low-rank updates to the quantized model for adaptation; QLoRA uses NF-quantization, while GPTQ-LoRA uses  approximate second-order information to solve for $\argmin_{\hat{\boldW} \in \mathbb{Q}_{b}^{d \times k}}\Vert \boldX\boldW - \boldX\ \hat{\boldW}\Vert_F$ \cite{frantar2022optimal,frantar-gptq}. As the original papers were applied on top of LLaMA-1 \cite{touvron2023llama}, for fair comparison we reimplement these baselines on top of LLaMA-2. We follow \citet{dettmers2023qlora} use rank = 64 for our main experiments, and ablate on the rank in our analysis section.

\vspace{-2mm}
\paragraph{Evaluation.}
To evaluate models trained on C4, we use three metrics: perplexity on C4 validation, perplexity on WikiText-2~\cite{merity2016pointer}, and 5-shot MMLU accuracy~\cite{hendrycks2021measuring}.  For instruction-tuned models,\footnote{We did not include GPTQ-LoRA in instruction-tuning experiments because the training was unstable.} we use a Vicuna-style automatic evaluation~\cite{vicuna2023open}. This involves asking GPT-$4$ to make pairwise comparisons between its outputs and those of GPT-$3.5$ (with the possibility of a tie) over $80$ curated questions. We chose this evaluation scheme over the $10$-point rating system, following the recommended setup from~\citet{dettmers2023qlora}.\footnote{However we do not use an ELO-style rating system (which would require evaluations across all possible pairs) due to the large number of models involved.} For the GLUE benchmark, we show the average metrics across all tasks. 

\begin{figure}[t]
\centering
\vspace{-14mm}
\includegraphics[width=.99\textwidth]{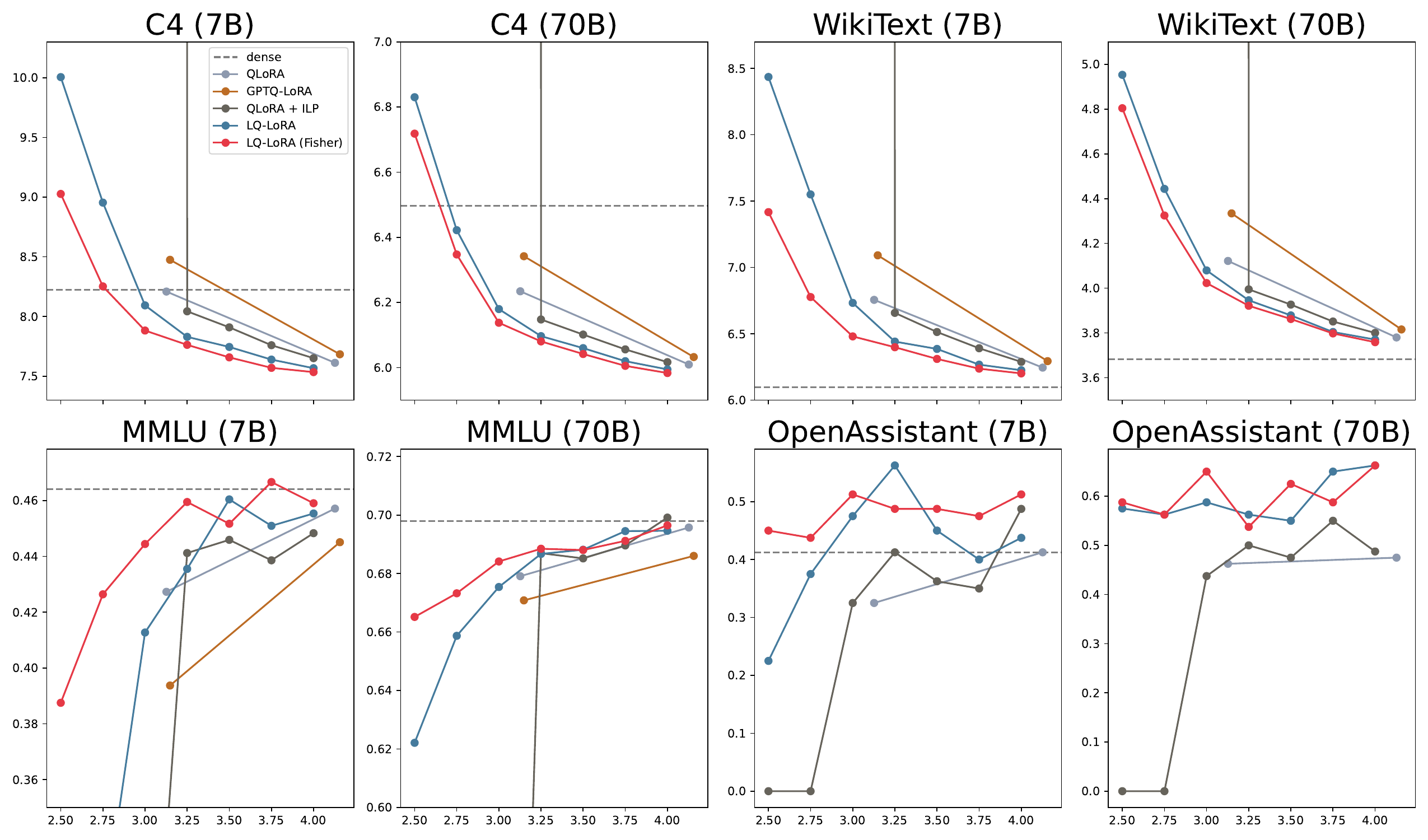}
\vspace{-3mm}
\caption{LQ-LoRA LLaMA-2 models with  rank = 64. C4/Wikipedia/MMLU results are based on finetuning on C4. Vicuna eval is based on finetuning on the OpenAssistant dataset. QLoRA \citep{dettmers2023qlora} and GPTQ-LoRA \citep{chai2023int2} are based on our own reimplementations. Dense refers to unquantized models (no training) except for instruction tuning experiments. In the latter case, dense refers to full finetuning (7B model only, OOM for 70B).}
\label{fig:llama2}
\vspace{-5mm}
\end{figure}

\vspace{-2mm}
\paragraph{Training details.}
Unless specified otherwise, we use a rank of $64$, no LoRA dropout, and a default learning rate of $2\times10^{-5}$, with a few exceptions. 
For continual language modeling, we train on one partition of the C4 data for half an epoch, using a sequence length of $1024$ for both training and evaluation.
To estimate the Fisher, we use $10000$ samples from C4 with a sequence length of $1024$. For the GLUE tasks, we use a similar setup, but with masked language modeling objectives on C4.
For instruction tuning, we use the hyperparameters suggested by~\citet{dettmers2023qlora} (except LoRA dropout). For GLUE fine-tuning, we follow the learning rate and number of epochs recommended by~\citet{hu2022lora} for the QLoRA baseline. However, we only fine-tune the model for $5$ epochs for MNLI and QQP due to their sizes.

\begin{wraptable}{r}{0.34\textwidth}
\vspace{-15mm}
\centering
\footnotesize
\begin{tabular}{lll}
\toprule
 Method & Bits & GLUE  \\
\midrule
Full FT & 16 & 88.5 \\
QLoRA 3-bit & 3.127 & 86.1 \\
\midrule
QLoRA  & 2.5 & 75.4  \\
(ILP) & 2.75 & 80.7 \\
  & 3.0 & 85.5 \\
   & 3.25 &  86.1 \\
   \midrule
   LQ-LoRA & 2.5 & 85.7 \\
 & 2.75 & 87.1 \\
  & 3.0 & 87.3 \\
   & 3.25 & 88.1 \\
   \midrule
   LQ-LoRA & 2.5 & 87.3 \\
(Fisher)  & 2.75 & 86.4 \\
  & 3.0 & 87.3 \\
   & 3.25 & 88.3 \\
\bottomrule
\end{tabular}
\vspace{-2.5mm}
\caption{Performance on GLUE with RoBERTa-Large.}
\label{tab:glue}
\vspace{-4mm}
\end{wraptable}

\vspace{-2mm}
\subsection{Results}
\vspace{-3mm}
Figure~\ref{fig:llama2} shows the results of language modeling and instruction tuning on LLaMA-2 across different model sizes and metrics. The full numeric results in Table~\ref{table:llama2} of Appendix~\ref{subsec:full-results}. In general we find that LQ-LoRA almost always outperforms QLoRA and GPTQ-LoRA at (near) similar bit budgets. For example, 3.5 bit (Fisher) LQ-LoRA is generally comparable to NF-4-bit QLoRA (which requires 4.127 bits/param); similarly, 2.75-bit LQ-LoRA is competitive with NF-3-bit QLoRA (which requires 3.127 bits/param). These comparisons highlight the utility of the mixed-quantization scheme since these mixed strategies would not even have been found without the ILP. It should be noted, however, that as we approach the $2.5$-bit range, performance begins to degrade significantly. At the smaller 7B scale, the Fisher-weighted version of LQ-LoRA outperforms the unweighted version by a significant margin at all target bit widths. However, this discrepancy shrinks at the 70B scale.

Table~\ref{tab:glue} shows GLUE benchmark  with RoBERTa-Large, where we observe similar trends: LQ-LoRA outperforms QLoRA at similar bit-widths, and Fisher-weighted LQ-LoRA is especially effective at 2.5 bits. 

\vspace{-2mm}
\subsection{LQ-LoRA for Model Compression}

\input{tables/results-ptq2}

\vspace{-2mm}
\paragraph{Results.}
Table~\ref{table:ptq} shows the results on C4 and WikiText, where we  follow prior PTQ works~\cite{frantar-gptq,shao2023omniquant} and measure performance through C4 and WikiText-2 perplexity on a specific subset of data. LQ-LoRA with 2.75 bits  results in an average bits/param of 2.95 bits and 2.85 bits for the 7B and 70B models respectively, when taking into account the LoRA components (``Effective bits'' in Table~\ref{table:ptq}). We find that this generally outperforms other sub-4-bit PTQ methods which also use calibration data to quantize the pretrained models.

Given the promising results with LQ-LoRA on continual language modeling, we next experiment with whether larger-scale language modeling can improve results further and enable the use of LQ-LoRA as a viable technique for model compression. 
Specifically, we take LQ-LoRA (Fisher, $2.75$-bits, $64$-rank) and fine-tune it on a larger calibration dataset of two C4 partitions and WikiText-2, using a sequence length of $2048$. We further quantize the low-rank components themselves using NF-$8$ configuration after training.\footnote{NF-8 replaces the first-level quantization of the original NF-4 with $8$-bits.} 

In Table~\ref{tab:openllm}, we evaluate the zero/few-shot capabilities using the Eleuther AI Language Model Evaluation Harness~\citep{evalharness}, a unified framework to test generative language models on a large number of different evaluation tasks. Specifically, we follow HuggingFace's the Open LLM Leaderboard\footnote{\url{https://huggingface.co/spaces/HuggingFaceH4/open_llm_leaderboard}} and evaluate models on 6 key benchmarks: ARC~\citep{clark2018think}, HellaSwag~\cite{zellers2019hellaswag}, MMLU~\cite{hendrycks2020measuring}, TruthfulQA~\cite{lin2022truthfulqa}, Winogrande~\cite{sakaguchi2021winogrande}, and GSM8k~\cite{cobbe2021training}.
We observe that there is nontrivial  degradation on some benchmarks (GSM8K, ARC), indicating that perplexity degradations are not always commensurate with downstream zero/few-shot performance.

\input{tables/results-lmeval}

\begin{wrapfigure}{r}{0.45\textwidth}
\centering
\vspace{-10mm}
\includegraphics[width=.45\textwidth]{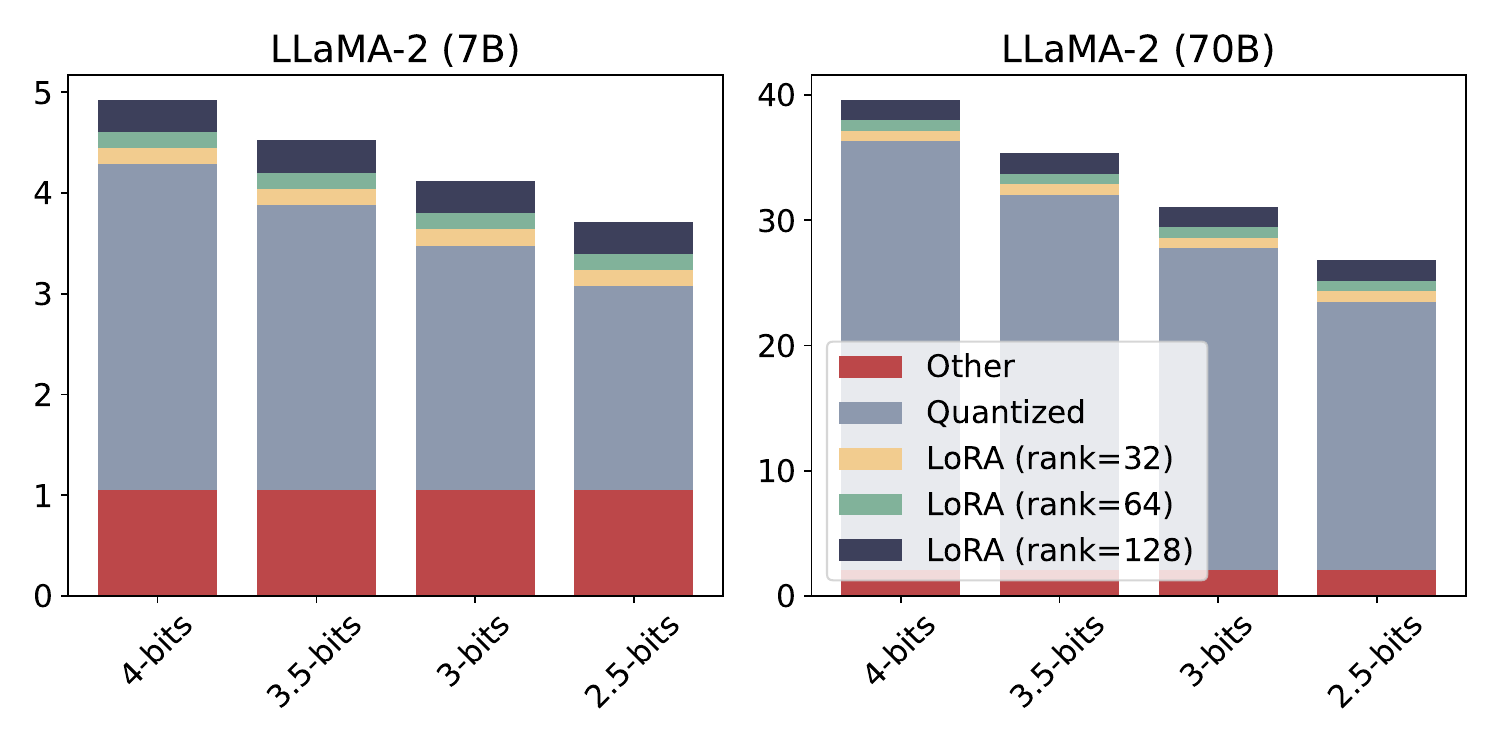}
\vspace{-7.9mm}
\caption{Storage in GB (y-axis) vs. bits (x-axis), broken down by quantized parameters, LoRA parameters, and others (e.g., embeddings). LLaMA-2 7B and 70B with 16-bits requires 14GB and 139GB, respectively.}
\label{fig:storage}
\vspace{-6mm}
\end{wrapfigure}

\vspace{-3mm}
\subsection{Analysis}
\begin{figure}[t]
\centering
\includegraphics[width=.99\textwidth]{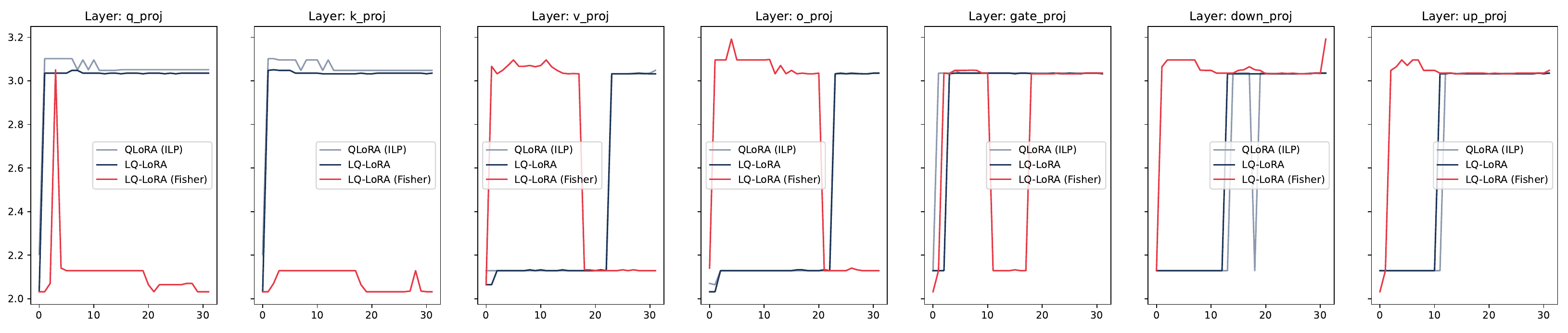}
 \vspace{-3mm}
\caption{Visualization of the bits/param allocated by ILP broken down by matrix type. The y-axis is the bits/param, while x-axis indicates layer number. We show the allocation for QLoRA, LQ-LoRA  and Fisher-weighted LQ-LoRA for target bit rate of 2.75 bits.}
\label{fig:allocations}
\vspace{-4mm}
\end{figure}

\vspace{-2mm}
\label{subsec:analysis}
\paragraph{Mixed-configuration quantization.}
We show the allocations of quantization configuration, measured by the average bits per parameter for a given matrix, in Figure~\ref{fig:allocations}. Each plot displays the decisions of the ILP for $2.75$ target bit rate. 
 ILP is able to  allocate different configurations to different matrices, and this decision is indeed different between  Fisher-weighted and  non-Fisher-weighted variants.

\vspace{-2mm}
\paragraph{LoRA ranks.} We investigate the effect of LoRA rank with the LLaMA-2-7b model in Table~\ref{table:rank-analysis} fixing the quantization configuration to NF-3 (i.e., 3.127 bits/param). QLoRA is  insensitive to the LoRA rank. However, LQ-LoRA is able to make ``better'' use of the additional ranks to minimize the error at initialization,  leading to improved  performance.
\vspace{-2mm}

\paragraph{Memory requirements.} In Figure~\ref{fig:storage} we show the memory required for storing the model for  different bit rates, broken down by the non-quantized component, the quantized component, and the LoRA parameters. Quantization into sub-3-bits greatly decreases the memory required for running the model. At sub-3 bits, it becomes possible to run the 70B model on a single GPU with 40GBs. Finetuning requires more memory due to memory required for the activations and LoRA gradients/optimizer states. However, we are able to run  full forward/backward passes on the sub-3-bit 70B models on a single 80GB GPU with batch size 2 and sequence length 2048.

%% file: tables/results-ptq2.tex
\begin{table}[t]
\vspace{-10mm}
\centering
\caption{LQ-LoRA comparison against other sub-4-bit PTQ methods. While we only experiment with LQ-LoRA on LLaMA-2 (bottom), we show other PTQ results on LLaMA-1 (top) as well to calibrate our results, as most prior works have focused on LLaMA-1.   ``Effective bits'' takes into account the extra storage needed to store quantization parameters (e.g., scaling factors). In LQ-LoRA this includes the LoRA components, which are themselves quantized to 8 bits. For other methods, we take results corresponding to a setting with 3-bit quantization and a group-size 128 (if possible, otherwise the closest one). The effective bits for SpQR and LQ-LoRA are dependent on model size, and hence we show the effective bits for both settings.
$^\dagger$Results from~\citet{shao2023omniquant}. $^\ddagger$We show  3.1-bits  instead of 3.01-bits with group size  128 because the latter performed worse.
}
\vspace{-2mm}
{\fontsize{8.5}{9}\selectfont
\begin{tabular}{ll|rr|rr}
\toprule
    \textbf{Method} &
    \textbf{Effective Bits} &
    \multicolumn{2}{r|}{\textbf{C4}} &
    \multicolumn{2}{r} {\textbf{WikiText}} \\
&
    (7B, 65B/70B) &
     7B &
    65B/70B &
     7B &
    65B/70B \\
\midrule
\textit{LLaMA-1 Uncompressed}$^\dagger$                                                          & \textit{16}          & \textit{7.08} & \textit{5.62} & \textit{5.68} & \textit{3.53} \\
\hspace{1mm} SpQR~\cite{dettmers2023spqr}                                             & 3.94, 3.90  & 7.28 & 5.70 & 5.87 & 3.68 \\

 \hspace{1mm} RTN~(\texttt{3-bits, g128})$^\dagger$                                    & 3.15        & 8.62 & 6.10 & 7.01 & 4.24 \\
 \hspace{1mm} GPTQ (\texttt{3-bits, g128})~\citep{frantar-gptq}$^\dagger$            & 3.15        & 7.85 & 6.00 & 6.55 & 4.17 \\
 \hspace{1mm} AWQ (\texttt{3-bits, g128})~\citep{lin2023awq}$^\dagger$               & 3.15        & 7.92 & 5.94 & 6.46 & 3.99 \\

 \hspace{1mm} PEQA (\texttt{3-bits, g128})~\citep{kim2023memory}                             & 3.15        & -    & -    & 5.91 & -    \\
 \hspace{1mm} OWQ (\texttt{3-bits})~\cite{lee2023owq}$^\ddagger$                                         & 3.1         & 8.15 & 6.16 & 6.39 & 4.08 \\
 \hspace{1mm} SqueezeLLM (\texttt{3-bits, 0.45\%})~\citep{kim2023squeezellm}                 & 3.24        & 7.56 & -    & 6.13 & -    \\
 \hspace{1mm} SqueezeLLM (\texttt{3-bits})~\citep{kim2023squeezellm}                                      & 3.02        & 7.75 & -    & 6.32 & -    \\
 \hspace{1mm} OmniQuant (\texttt{3-bits, g128})~\citep{shao2023omniquant}$^\dagger$  & 3.15        & 7.75 & 5.93 & 6.15 & 3.94 \\
 \hspace{1mm} OmniQuant (\texttt{2-bits, g64})~\citep{shao2023omniquant}$^\dagger$     & 2.28       & 11.78 & 7.60 & 8.90 & 5.65 \\
 \hspace{1mm} LREC (\texttt{3-bits, g128})~\cite{chai2023int2} & 3.35 &  8.24 & - & 5.52 & - \\
 \hspace{1mm} LREC (\texttt{2-bits, g128})~\cite{chai2023int2} & 2.24 & 12.52 & - & 8.74 & - \\
\midrule
 \textit{LLaMA-2 Uncompressed}$^\dagger$                                                          & \textit{16}         & \textit{6.97} & \textit{5.52} & \textit{5.47} & \textit{3.31} \\
\hspace{1mm} RTN~(\texttt{3-bits, g128})$^\dagger$                                    & 3.15        & 8.40 & 6.02 & 6.66 & 3.97 \\
 \hspace{1mm} GPTQ (\texttt{3-bits, g128})~\citep{frantar-gptq}$^\dagger$            & 3.15        & 7.89 & \textbf{5.85} & 6.29 & 3.85 \\
 \hspace{1mm} AWQ (\texttt{3-bits, g128})~\citep{lin2023awq}$^\dagger$               & 3.15        & 7.84 & -    & 6.24 & -    \\
 \hspace{1mm} OmniQuant (\texttt{3-bits, g128})~\citep{shao2023omniquant}$^\dagger$  & 3.15        & 7.75 & \textbf{5.85} & 6.03 & 3.78 \\
  \hspace{1mm} OmniQuant (\texttt{2-bits, g64})~\citep{shao2023omniquant}$^\dagger$   & 2.28       & 12.72 & 7.88 & 9.62 & 6.11 \\

 \hspace{1mm} LQ-LoRA (\texttt{2.75{-}bits, 64{-}rank, Fisher})   & 2.95, 2.85 & \textbf{7.60} & {5.88} & \textbf{5.67} & \textbf{3.65} \\
\bottomrule
\end{tabular}
}
\vspace{-4mm}
\label{table:ptq}
\end{table}

%% file: tables/results-lmeval.tex
\begin{table}
\small
\centering
{\fontsize{7.75}{9}\selectfont
\caption{Performance on HuggingFace's Open LLM benchmark with LLaMA-2. We use the same LQ-LoRA setup as in Table~\ref{table:ptq} (\texttt{2.75{-}bits, 64{-}rank, Fisher}).}
\vspace{-3mm}
\label{tab:openllm}
\begin{tabular}{ll|llllll|l}
\toprule
Method & Size & ARC & HellaSwag & MMLU & TruthfulQA & Winogrande & GSM8K & \textbf{Average} \\
\midrule
Uncompressed (16 bits)  & 7B & 53.2 & 78.6 & 39.0 & 46.6 & 73.6 & 14.9 & 51.0 \\
LQ-LoRA (2.95 bits) & 7B & 49.8 & 75.9 & 39.3 & 43.0 & 72.4 &  7.4 & 48.0 \\
\midrule
Uncompressed (16 bits)   & 70B & 67.2 & 87.3 & 44.8 & 69.6 & 83.7 & 53.7 & 67.7 \\
LQ-LoRA (2.85 bits) & 70B & 65.8 & 86.2 & 44.5 & 66.9 & 83.2 & 45.6 & 65.3 \\
\bottomrule
\end{tabular}
}
\vspace{-3mm}
\end{table}

%% file: discussions.tex
\vspace{-2mm}
\section{Discussion and Limitations}
\vspace{-2mm}
\label{sec:discussion}
Our simple iterative algorithm was found to be empirically effective but is ultimately heuristic, and it would be interesting to see if more theoretically-principled optimization algorithms could be derived. And while we focused on NF-quantization to enable comparison against QLoRA, applying LQ decomposition on top of other quantization approaches could result in further gains. It is also possible to extend the ILP-based mixed-precision approach to mixed-precision quantization \emph{and} mixed-rank decomposition to enable the assignment of different ranks to each matrix.\footnote{However, this may be suboptimal since the ILP only minimizes decomposition error, and not the final downstream performance. This could be addressed by weighting the LoRA parameters as less costly (since the finetunable  parameters contribute more for downstream performance) in the ILP.}

We also discuss some negative results as well as limitations of LQ-LoRA.  We found that re-factorizing the matrix periodically (e.g., after every $K$ gradient steps) did not yield improvements. Insofar as our initialization of $\boldL_1$ and $\boldL_2$ could orient the model to adapt itself in ways that may not be optimal for the task at hand, we also tried a hybrid approach where half of the adaptable  low-rank component comes from LQ-LoRA, and the other half comes from standard LoRA initialization, but did not find this to improve results. Our approach also heavily relies on the fact that adaptation will occur through low-rank updates, and thus is not generally applicable to other parameter-efficient finetuning methods.

\input{tables/results-ranks}

%% file: tables/results-ranks.tex
\begin{table}
\vspace{-10mm}
\small
\centering
\caption{C4 and WikiText perplexity as a function of LoRA ranks. For both QLoRA and LQ-LoRA we used fixed NF-3 configuration for all layers, which incurs an average cost of 3.127 bits/param. The rightmost column shows the sum of errors across all matrices in LLaMA-2-7b, which corresponds to  $\left\Vert \boldW - \quantize(\boldW)\right\Vert_F^2$ for QLoRA and  $\left\Vert \boldW - \left(\boldQ + \boldL_1 \boldL_2\right) \right\Vert_{F}^2$ for LQ-LoRA.}
\vspace{-2mm}
\begin{tabular}{ll|rr|r}
\toprule
\textbf{Method} & \textbf{LoRA rank} & \textbf{C4} & \textbf{WikiText} & \textbf{Error} \\
\midrule
{QLoRA 3-bit}
 & 32  & 8.21 & 6.75  &  \\
{(3.127 bits/param)} & 64  & 8.21 & 6.76  & $9.83 \times 10^4$ \\
 & 128 & 8.21 & 6.76  &  \\
\midrule
{LQ-LoRA 3-bit}
& 32  & 8.02  & 6.61 & $7.99 \times 10^4$ \\
{(3.127 bits/param)}  & 64  & 7.93  & 6.51 & $7.12 \times 10^4$ \\
 & 128 & 7.84  & 6.46 & $5.98 \times 10^4$ \\
\bottomrule
\end{tabular}
\vspace{-6mm}
\label{table:rank-analysis}
\end{table}

%% file: related-works.tex
\vspace{-2mm}
\section{Related Work}
\vspace{-2mm}
\paragraph{Parameter-efficient finetuning.} Our work is closely related to parameter-efficient finetuning. Popular methods include Adapters~\citep{houlsby2019parameter,mahabadi2021parameter}, which insert trainable layers, prompt tuning~\cite{li-liang-2021-prefix,lester-etal-2021-power}, which optimizes continuous prompts,  and other methods which update subparts of the parameter vector~\cite{guo2021diff,zaken2021bitfit,NEURIPS2021_cb2653f5,hu2022lora}. (Some) parameter-efficient finetuning methods can reduce the GPU memory required for finetuning as there is no need to store the optimizer states associated with the fixed parameters. Recent work also combines parameter-efficient finetuning methods with quantization \cite{kwon-etal-2022-alphatuning,dettmers2023qlora}.
 
\paragraph{Low-rank plus sparse/quantized matrix decomposition.} Decomposing a data matrix into a low-rank matrix plus a sparse matrix (also known as robust PCA) is  well-studied  from both theoretical and applied perspectives \cite[][\textit{inter alia}]{lin2010augmented,zhou2011godec,pmlr-v31-zhou13b,liu2013robust,aravkin2014,hinter2014robust,yi2016fast,zhang2017robust}. Within deep learning robust PCA has previously been applied to compress smaller models with fewer than 100M parameters \cite{chen2018deep,cai2021learned}. \citet{saha2023matrix} uses sketching techniques to obtain a quantized, low-rank approximation of a pretrained matrix. Recent contemporaneous work \cite{li2023loftq} also performs low-rank plus quantized decomposition for LLM adaptation.

\paragraph{LLM compression.} While there has been much work on low-rank compression of smaller LLMs with fewer than 1B parameters  \cite{chen2021drone,tukan2021robust,tahaei2021kroneckerbert}, low-rank approaches for 1B+ LLMs remain underexplored, possibly because singular values of the pretrained matrices of LLMs have been found to  decay slowly \cite{chen2021drone}. Existing approaches for LLM compression have thus generally focused on quantization. Much recent work has focused on data-aware quantization strategies \cite{dettmers2022llm8,xiao2022smoothquant,dettmers2023spqr,frantar-gptq,kim2023squeezellm,lin2023awq}.

%% file: conclusion.tex
\vspace{-2mm}
\section{Conclusion}
\vspace{-2mm}
This work proposes a simple extension of LoRA which factorizes the pretrained matrices into low-rank and quantized components, where the quantization component can employ a dynamic configuration strategy. We observed this low-rank plus quantized decomposition approach to yield meaningful improvements over strong baselines.

%% file: appendix.tex
\newpage

\section{Implementation Details}
\label{subsec:implementation}
Here we discuss some  implementation details for efficiently implementing LQ-LoRA.

\begin{figure}[t]
\centering
\begin{subfigure}[b]{1.0\textwidth}
   \includegraphics[width=1\linewidth]{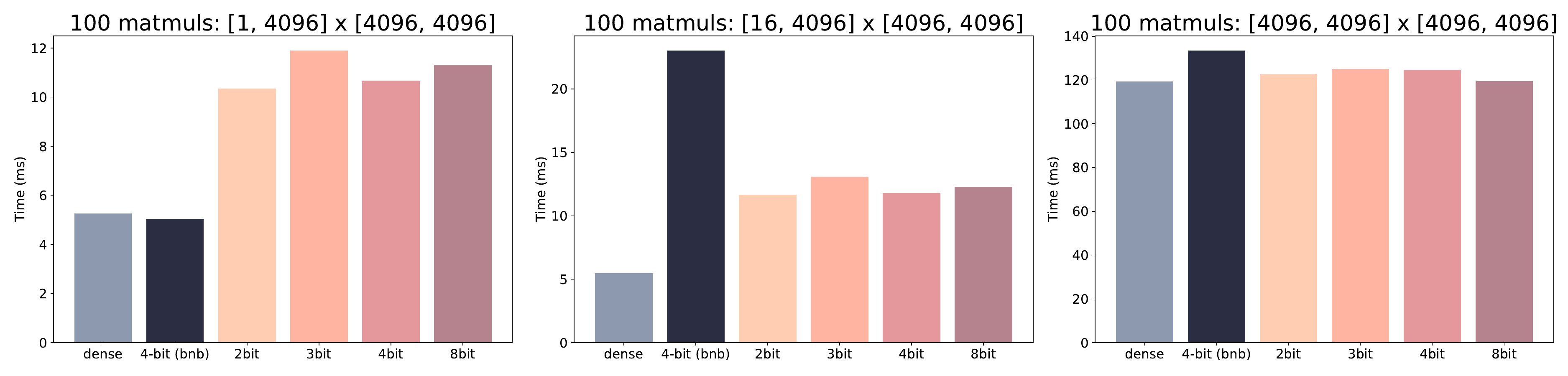}
   \caption{Run-times on A100 GPU}
\end{subfigure}
\begin{subfigure}[b]{1.0\textwidth}
   \includegraphics[width=1\linewidth]{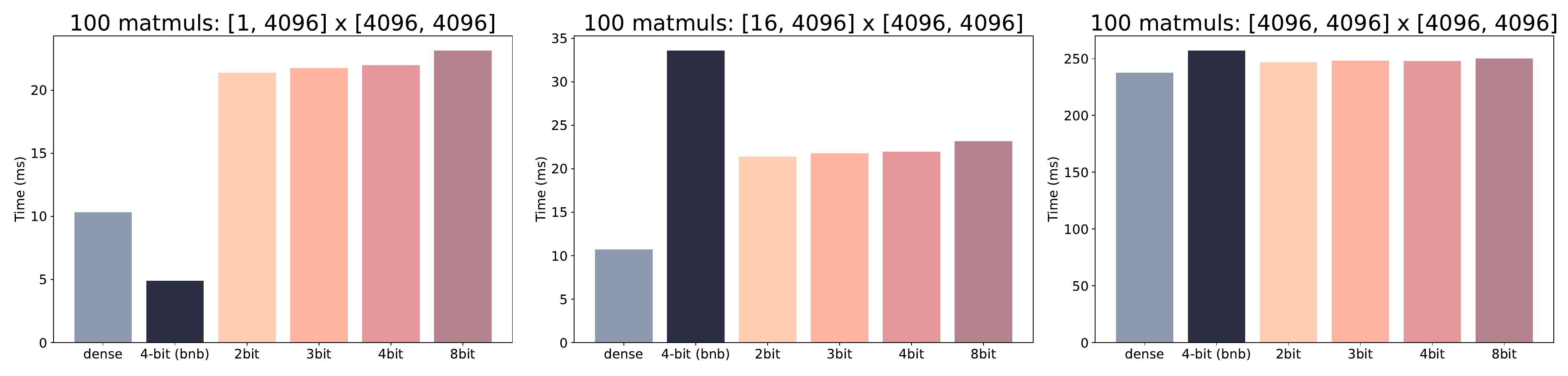}
   \caption{Run-times on A6000 GPU}
\end{subfigure}
\caption{
A100/A6000 GPU runtime to perform $100$ matrix-matrix multiplications in \texttt{fp32} between input data and quantized matrices (which involves on the fly dequantization). \texttt{bitsandbytes} (\texttt{bnb})~\cite{dettmers2023qlora} has separate implementations for training and for inference (matrix-vector multiplications, leftmost figure). We use the same quantization configuration as NF-$4$ and vary the first-level bits ($2,3,4,8$) for consistent comparisons.
}
\label{fig:runtimes}
\end{figure}

\paragraph{PyTorch-based mixed-quantization.}
Weight-only quantization techniques typically require packing sub-8-bit matrices into natively-supported data types (e.g., \texttt{int8}), and then unpacking to float-point format during dequantization.  As such, existing implementations often require custom CUDA extensions that are dependent on a particular quantization configuration, making it difficult to extend to mixed-quantization strategies. Our implementation is based entirely on PyTorch for fast experimentation and implementation of dynamic quantization strategies. We use PyTorch’s \texttt{\_\_torch\_dispatch\_\_} functionality to duck-type \texttt{torch.Tensor},\footnote{\url{https://dev-discuss.pytorch.org/t/what-and-why-is-torch-dispatch/557}}  which redefines behaviors under PyTorch operations such as addition and matrix multiplication. We then use PyTorch’s (full-graph) compiler to compile the following operations: (1) bits-unpacking, (2) dequantization, (3) linear algebra operations such as \texttt{add} and \texttt{matmul}, and (4) transpose and casting (for \texttt{bf16} training). For LoRA finetuning, we observed this PyTorch-based implementation (followed by compilation) to be as fast as QLoRA's \texttt{bitsandbytes} implementation,\footnote{\url{https://github.com/TimDettmers/bitsandbytes}} which relies heavily on CUDA extensions that are tailored for  4-bit NF quantization.

\paragraph{LoRA optimizer offloading.}
We also optionally work with  a CPU-based optimizer~\citep{ren2021zero}, which extends the pageable optimizer proposed in~\citet{dettmers2023qlora}. This implementation takes advantage of the fact that in LoRA, only a small portion of parameters needs to be trained, which makes data movement between CPU and GPU, as well as computation on CPU, relatively manageable. We retain a copy of trainable parameters on CPU, offload gradients from GPU to CPU before executing optimizer step on the parameter copy on CPU, and copy them back into GPU. We overlap the per-matrix optimizer step and CPU to GPU movement through async copy. On the largest 70 billion parameter model, we noticed a $14$\% memory saving with only a marginal (${<}2$\%) increase in training speed with this strategy.

\paragraph{Runtimes.}
\label{paragraph:runtimes}
Figure~\ref{fig:runtimes} displays the run-times of matrix multiplications between \texttt{FP32} input data and quantized matrices. We dequantize the matrix just-in-time before executing the \texttt{matmul}, and hence the runtime is lower-bounded by \texttt{FP32} \texttt{matmul} (dense). We enable \texttt{TF32} in PyTorch to utilize Tensor Cores, and collect CUDA time through PyTorch Profiler. Notably, the dequantization overhead is relatively small for reasonably-sized matrices.

\section{Full Results}
Table~\ref{table:llama2} shows the full numeric results of LQ-LoRA vs. QLoRA and GPTQ-LoRA.
\label{subsec:full-results}
\input{tables/results-lm-it}

%% file: tables/results-lm-it.tex
\begin{table}
\vspace{-8mm}
\footnotesize
\centering
\caption{LQ-LoRA LLaMA-2 models with  rank = 64. C4/Wikipedia/MMLU results are based on finetuning on C4. Vicuna eval is based on finetuning on the OpenAssistant dataset. QLoRA \citep{dettmers2023qlora} and GPTQ-LoRA are based on our own reimplementations.}
\vspace{-2mm}
\begin{tabular}{ll|rr|rr|rr|rr}
\toprule
\textbf{Method} & \textbf{Bits per} & \multicolumn{2}{r|}{\textbf{C4 (PPL) }} & \multicolumn{2}{r|}{\textbf{WikiText (PPL)}} & \multicolumn{2}{r|}{\textbf{MMLU (acc.)}} & \multicolumn{2}{r}{\textbf{Vicuna Eval}} \\
 & \textbf{param}  & 70B & 7B & 70B & 7B & 70B & 7B & 70B & 7B \\
\midrule
Dense (no training) & - & 6.50 & 8.22 & 3.68 & 6.10 & 0.70 & 0.46  & - & - \\ %
Dense (full finetuning) & - & - & - & - & - & - & - & OOM & 0.41 \\
\midrule
 QLoRA 3-bit & 3.127 & 6.23 & 8.21 & 4.12 & 6.76 & 0.68 & 0.43 & 0.46 & 0.33 \\ %
 QLoRA 4-bit & 4.127 & 6.01 & 7.61 & 3.78 & 6.25 & 0.70 & 0.46 & 0.47 & 0.41 \\ %
\midrule
 GPTQ-LoRA 3-bit & 3.148  & 6.34 & 8.48 & 4.33 & 7.09 & 0.67 & 0.39 & - & - \\ %
 GPTQ-LoRA 4-bit & 4.156  & 6.03 & 7.68 & 3.82 & 6.29 & 0.69 & 0.45 & - & - \\ %
\midrule
\multirow[t]{7}{*}{QLoRA + ILP} & 2.50 & 2223.2 & 2996.3 & 3319.4 & 4084.3 & 0.23 & 0.23 & 0.00 & 0.00 \\ %
 & 2.75 & 2193.9 & 2736.5 & 3292.6 & 3932.2 & 0.23 & 0.27 & 0.00 & 0.00 \\ %
 & 3.00 & 1781.5 & 1969.3 & 2587.0 & 3091.0 & 0.23 & 0.23 & 0.44 & 0.33 \\ %
 & 3.25 & 6.15 & 8.04 & 3.99 & 6.66 & 0.69 & 0.44 & 0.50 & 0.41 \\ %
 & 3.50 & 6.10 & 7.91 & 3.93 & 6.51 & 0.69 & 0.45 & 0.47 & 0.36 \\ %
 & 3.75 & 6.06 & 7.76 & 3.85 & 6.39 & 0.69 & 0.44 & 0.55 & 0.35 \\ %
 & 4.00 & 6.02 & 7.65 & 3.80 & 6.29 & 0.70 & 0.45 & 0.49 & 0.49 \\ %
\midrule
\multirow[t]{7}{*}{LQ-LoRA} & 2.50 & 6.83 & 10.00 & 4.95 & 8.44 & 0.62 & 0.31 & 0.57 & 0.23 \\ %
 & 2.75 & 6.42 & 8.95 & 4.44 & 7.55 & 0.66 & 0.31 & 0.56 & 0.38 \\ %
 & 3.00 & 6.18 & 8.09 & 4.08 & 6.73 & 0.68 & 0.41 & 0.59 & 0.47 \\ %
 & 3.25 & 6.10 & 7.83 & 3.95 & 6.44 & 0.69 & 0.44 & 0.56 & 0.56 \\ %
 & 3.50 & 6.06 & 7.75 & 3.88 & 6.39 & 0.69 & 0.46 & 0.55 & 0.45 \\ %
 & 3.75 & 6.02 & 7.64 & 3.80 & 6.27 & 0.69 & 0.45 & 0.65 & 0.40 \\ %
 & 4.00 & 5.99 & 7.57 & 3.77 & 6.23 & 0.69 & 0.46 & 0.66 & 0.44 \\ %
\midrule
\multirow[t]{7}{*}{LQ-LoRA (Fisher)} & 2.50 & 6.72 & 9.03 & 4.80 & 7.42 & 0.67 & 0.39 & 0.59 & 0.45 \\ %
 & 2.75 & 6.35 & 8.25 & 4.32 & 6.78 & 0.67 & 0.43 & 0.56 & 0.44 \\ %
 & 3.00 & 6.14 & 7.88 & 4.02 & 6.48 & 0.68 & 0.44 & 0.65 & 0.51 \\ %
 & 3.25 & 6.08 & 7.76 & 3.92 & 6.40 & 0.69 & 0.46 & 0.54 & 0.49 \\ %
 & 3.50 & 6.04 & 7.66 & 3.86 & 6.31 & 0.69 & 0.45 & 0.62 & 0.49 \\ %
 & 3.75 & 6.01 & 7.57 & 3.80 & 6.24 & 0.69 & 0.47 & 0.59 & 0.47 \\ %
 & 4.00 & 5.98 & 7.53 & 3.76 & 6.20 & 0.70 & 0.46 & 0.66 & 0.51 \\ %
\bottomrule
\end{tabular}
\vspace{-4mm}
\label{table:llama2}
\end{table}